\documentclass{article}
\usepackage[style=numeric,sorting=none,maxbibnames=99]{biblatex}  
\usepackage{graphicx}
\usepackage{caption}
\usepackage{bm}
\usepackage{spconf}
\usepackage{booktabs}
\title{Mask Guided Attention For Fine-Grained Patchy Image  Classification}
\name{Jun Wang,\thanks{This work is done during Jun Wang's research internship at Griffith University.} Xiaohan Yu, Yongsheng Gao}
\address{Griffith University, QLD 4111, Australia \\
jun.1.wang@kcl.ac.uk, xiaohan.yu@griffith.edu.au, yongsheng.gao@griffith.edu.au}

\begin{document}
\ninept
\maketitle
\begin{abstract}
In this work, we present a novel mask guided attention (MGA) method for fine-grained patchy image classification. The key challenge of fine-grained patchy image classification lies in two folds, ultra-fine-grained inter-category variances among objects and very few data available for training. This motivates us to consider employing more useful supervision signal to train a discriminative model within limited training samples. Specifically, the proposed MGA integrates a pre-trained semantic segmentation model that produces auxiliary supervision signal, \textit{i.e.}, patchy attention mask, enabling a discriminative representation learning. The patchy attention mask drives the classifier to filter out the insignificant parts of images (\textit{e.g.}, common features between different categories), which enhances the robustness of MGA for the fine-grained patchy image classification. We verify the effectiveness of our method on three publicly available patchy image datasets. Experimental results demonstrate that our MGA method achieves superior performance on three datasets compared with the state-of-the-art methods. In addition, our ablation study shows that MGA improves the accuracy by 2.25\% and 2\% on the SoyCultivarVein and BtfPIS datasets, indicating its practicality towards solving the fine-grained patchy image classification.
\end{abstract}

Mask, attention, semantic segmentation, fine-grained patchy image classification
\endkeywords{}

\section{introduction}
Owing to the development of the hardware, large-scale datasets and well-designed deep learning algorithms, fine-grained image classification has obtained remarkable progress in past decades. However, fine-grained patchy image classification \cite{yu2020patchy}, which further moves down the granularity of categorization, \textit{e.g.} from species to cultivar, and covers fewer training data, remains an open challenge \cite{YU2021108067}.\par
Despite state-of-the-art methods achieving desirable performance on fine-grained classification tasks, these methods may not function well on fine-grained patchy image classification. For instance, to learn the most discriminative feature representations, a common strategy \cite{huang2016part,peng2017object}  is to utilize additional bounding box annotations to directly localize the most discriminative parts or foreground in images, and then use these localized regions for classification. Nonetheless, transferring these methods to fine-grained patchy image classification is difficult since patchy images barely suffer from the influence of other categories and cluttered background as shown in Figure 1.\par
Alternatively, attention mechanism \cite{xiao2015application,sun2018multi,zhao2017diversified} are utilized to automatically search the informative areas in images, alleviating the aforementioned problem. However, these models are trained with unsupervised manners, thus may be trapped into the local optimum. Especially for fine-grained patchy image classification, the unsupervised attention mechanism on such a small-scale dataset is highly likely to encounter the over-fitting problem. \par
To that end,  we propose a mask guided attention approach to effectively localize the discriminative local regions for fine-grained patchy image classification. Unlike previous attention mechanisms that run in an unsupervised manner, we utilize patchy mask as the ground truth of the attention map to supervise the learning of the automatically generated attention. In this way, the learned attention map is less likely to be trapped into local optimum and can accurately reflect the areas that to be focused on. In addition, the over-fitting problem can be alleviated since the mask-guided attention drives the classifier to focus on the informative and discriminative parts, instead of memorizing the whole images. \par
In this paper, we introduce a novel mask guided attention approach to effectively supervise the learning of attention map in fine-grained patchy image classification. The proposed method can be applied for those datasets that are not influenced significantly by clutter background, or the dicriminative regions can not be simply annotated by a bounding box. Our method obtains superior performance on three publicly available datasets. The ablation study demonstrates that MGA improves the accuracy by 2.25\% and 2\% on the SoyCultivarVein and the BtfPIS datasets against the attention mechanism without the mask supervision, verifying the effectiveness of our proposed method.

\begin{figure*}[ht]   
    \centering 
    \includegraphics[width=0.67\textwidth]{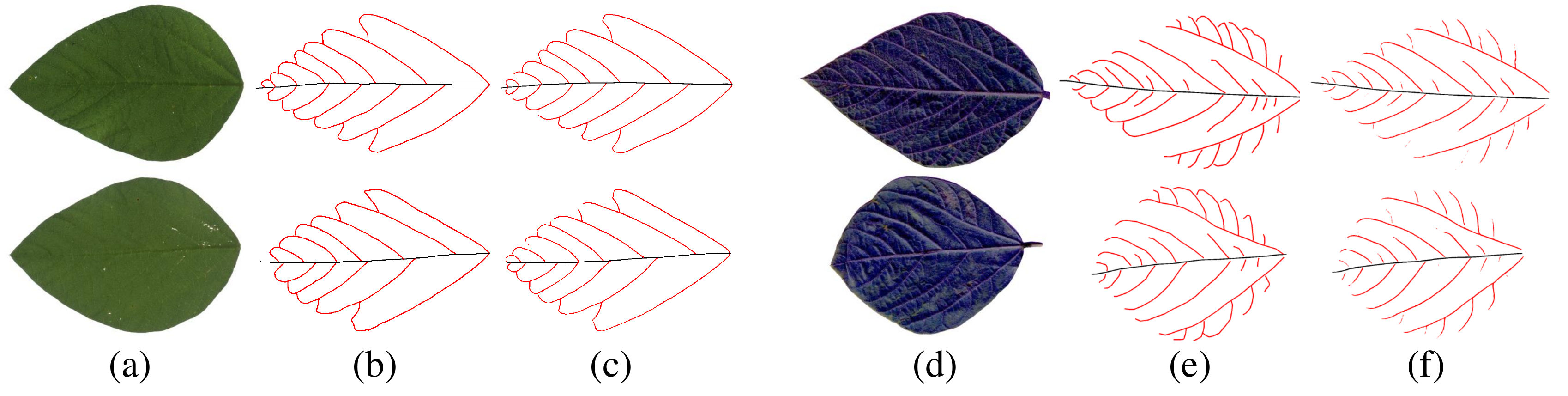} 
    \caption{Four examples of the training images in SoyCultivarVein (left part) and HainanLeaf (right part) datasets. (a) and (d) are the original training images coming from two species. (b) and (e) are their ground truth of two level masks corresponding to main vein (black) and second-order vein (red). (c) and (f) are the predicted masks.}    
    \label{fig_leaf_example} \vspace{-0.1cm}
\end{figure*}

\section{Related works}

\subsection{Fine-grained Patchy Image Classification}
Existing methods  \cite{yu2019contour,yu2020patchy,yu2015leaf,yu2016multiscale} for fine-grained patchy image classification rely on handcrafted feature representation, \textit{e.g.}, contour-based feature representations. Such feature representations, however, lack robustness and generalization capability, limiting their performance in fine-grained patchy image classification. To handle this problem, 
we propose to adopt the convolutional neural networks (CNN) as the feature extractor and directly learn the informative features in images, which is similar to existing fine-grained image classification settings.\par
Approaches in fine-grained classification can be coarsely divided into two groups. One \cite{huang2016part,peng2017object} is directly learning the features from discriminative parts of images by taking advantage of additional bounding boxes annotation of datasets. Similar to object detection, region proposals should be selected first, and then sent to the classifier to obtain the predicted categories. Although these methods perform well on datasets of which suffer from the background noise, limitation exists in those fine-grained patchy image datasets that the discriminative areas cannot be directly annotated by a rectangular. Another group of approaches try to automatically detect the discriminative areas via attention mechanism and have gained great success \cite{ZHAO2021107938}. \cite{xiao2015application} integrated three types of attention into the neural networks to localize the discriminative parts. \cite{sun2018multi} presented a new method to learn multi-attention region features and demonstrates substantial performance improvements on their proposed Dogs-in-the-Wild dataset. \cite{zhao2017diversified} developed a diversified visual attention network which significantly relieves the dependency on strongly supervised information. One drawback of these attention methods is that the generated attention map may be trapped into local optimum as they are designed in an unsupervised scenario. To cope with the aforementioned problems in the bounding boxes based-methods and attention-based methods, we propose to employ more informative signal, the patchy mask, to supervise the learning of attention by considering the mask as the ground truth of the attention map.
\subsection{Mask-guided Attention}  
Mask guided attention \cite{song2018mask}  has demonstrated improved performance in person re-identification tasks. \cite{pang2019mask,xie2020mask} further demonstrated the superiority of the mask-guided attention in detection tasks. To our knowledge, however, few study has explored the performance of mask-guided attention in fine-grained patchy image. Our work is motivated by \cite{song2018mask}, which adds mask as an additional channel in inputs and utilizes the mask-guided attention to learn full image, foreground image and background image features. These features are 
optimized by the usual triplet loss in the final 128 dimension features. Different from \cite{song2018mask} that having large-scale dataset available, this work focuses on patchy image datasets, on which training a classifier from scratch is of great difficulty if only few data is provided. Therefore, we can not simply add the mask as an additional channel in inputs, and exploit the mask to guide the learning of attention in the middle-level features. 

\begin{figure*}[ht]   
    \centering 
    \includegraphics[width=0.95\textwidth]{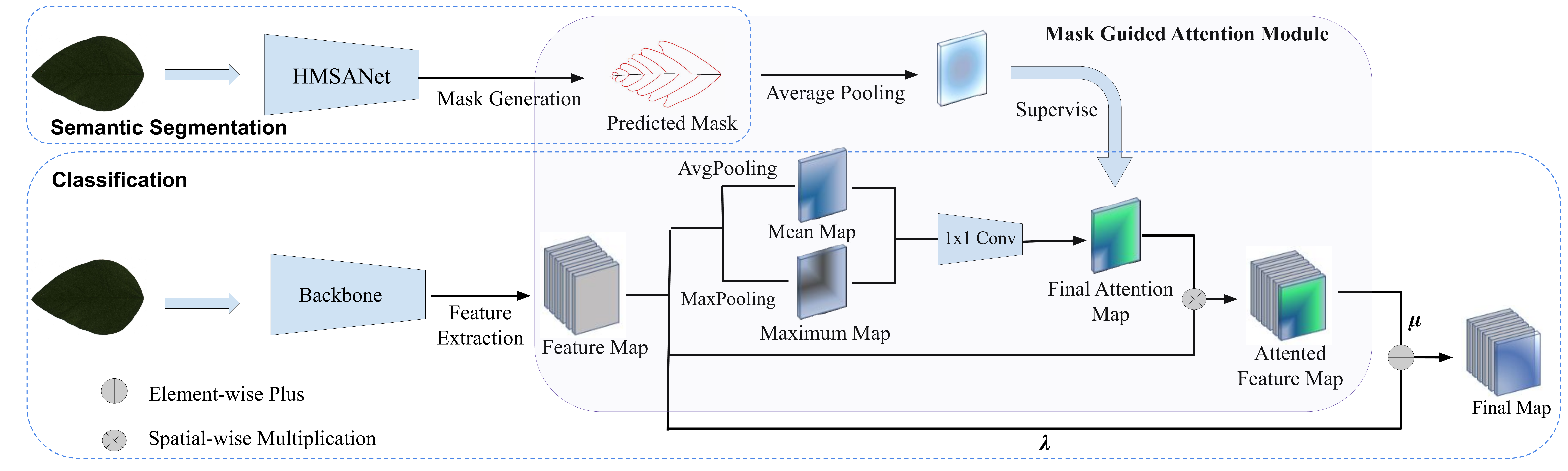} 
    \caption{The overall architecture of the proposed method which contains two parts. In the first part, a semantic segmentation model called HMSANet is trained to generate the mask of the training images. The mask is then resized to the same size of the final image feature map via average pooling. The second part undertakes the classification task. Given an input image, the associated feature is extracted by the backbone network and then fed into the Mask Guided Attention Module to generate the attention map. The resized mask is introduced here to supervise the learning of the final attention map.}    
    \label{fig_overall_architecture}  
\end{figure*}

\section{Method}

\subsection{Overall Architecture}
The overall architecture can be generally divided into two parts. In the first part, a semantic segmentation model is trained to obtain the predicted masks of images which are then considered as the ground truth for the attention map. This work selects HMSANet \cite{tao2020hierarchical} as the semantic segmentation model due to its state-of-the-art performance in semantic segmentation. Combined with the training images, those predicted masks are then fed into the second part, the classification part, to perform fine-grained patchy image classification.\par
As shown in Figure 2, the classification part first employs a CNN backbone to extract image features denoted as $ \bm{f}_{img}\subseteq\bm{R}^{H\times W \times C} $. Note that $H$ and $W$ are the image height and width, respectively, and $C$ is the number of the channel. An attention map $\bm{Am}\subseteq\bm{R}^{H\times W}$, generated by the mask-guided attention module, is then utilized to drive the feature extractor to recognize the informative and disriminative parts in images. Given the attention map $\bm{Am}$ and the image features $ \bm{f}_{img}$, the attended features $ \bm{f}_{att}\subseteq\bm{R}^{H\times W \times C} $ are then produced by:
\begin{equation}
 \bm{f}_{att}(i)=\bm{f}_{img}(i)\ast\bm{Am}, 
\end{equation}
where $i$ is the channel index in $\bm{f}_{img}$ and $\bm{f}_{att}$, $\ast$ is the spatial-wise multiplication. \par
It can be seen from Figures 1 and 2 that the aim of the semantic segmentation part is to segment the discriminative parts, e.g., the main vein and the second-order vein in leaves. However, this task is extremely challenging due to the ultra-fine segmentation granularity. In addition, insufficient data enhances the difficulty in training a general, robust and accurate semantic segmentation models. Under this setting, masks generated by the trained semantic segmentation model may contain much noise and inaccurate information. Hence, to alleviate the influence of the false positives, we take advantage of both the attended features $\bm{f}_{att}$ and the original image features $\bm{f}_{img}$ to obtain the final outputs.\par
Specifically, given the original features $\bm{f}_{img}$ and the attended features $\bm{f}_{att}$, the final features $\bm{f}_{final}$ sent to the fully connected layer for classification are calculated as :
\begin{equation}
 \bm{f}_{final}=\lambda\bm{f}_{img}+\mu\bm{f}_{att},\  
\end{equation}
\begin{equation}
s.t.\ \lambda+\mu=1,
\end{equation}
where $\lambda$ and $\mu$ are two hyper-parameters to balance the contribution between two features. Both of them are set to 0.5 in this paper.

\subsection{Mask-guided Attention Module}
In artificial neural networks, visual attention mechanism can be simply interpreted as a means of re-weighting the response of the feature maps, simulating the human visual system with the aim of enhancing the response of informative parts and reducing the activation in those trivial areas. The architecture of proposed MGA module is illustrated in the pink rectangular as shown in Figure 2. \par
In this paper, we follow the idea of \cite{woo2018cbam} to implement our spatial visual attention. To be specific, we do this by three steps. Firstly, given the image feature maps $\bm{f}_{img}$, average pooling and maximum pooling are applied to obtain the mean spatial attention map $\bm{Am}_{mn}$ and maximum spatial attention map $\bm{Am}_{max}$, as shown in Eqs.(4) to (6):
\begin{equation}
 \bm{Am}_{mn}^{i,j}=\frac{1}{C}\sum_{c=1}^{C}\bm{f}_{img}^{i,j}(c),  
\end{equation}
\begin{equation}
 \bm{Am}_{max}^{i,j}=\bm{f}_{img}^{i,j}(c*),
\end{equation}
in which, 
\begin{equation}
c*=argmax_c(\bm{f}_{img}^{i,j}(c)),
\end{equation}
\noindent
where $i,j$ are the coordinates in the attention map, and $c$ is the $c^{th}$ channel of the image feature maps.\par

Then, the mean and maximum attention maps are aggregated into the final one channel attention map $\bm{Am}$ via a $\bm{1\times 1}$ Conv layer. This allows the network to learn and update the weights between two attention maps, which can be regraded as a channel-wise attention. Sigmoid function is applied as the activation function, rendering the values of the attention map to fall into a bracket of [0,1]. Equation (7) shows this procedure, where $f_{1\times 1}$ is the $\bm{1\times 1}$ Conv layer, $\sigma$ is the sigmoid function. 
\begin{equation}
 \bm{Am}=\sigma(f_{1\times 1}(\bm{Am}_{mn};\bm{Am}_{max})).
\end{equation}

As stated in the previous section, attention usually runs in an unsupervised manner, hence global optimum is not always guaranteed. Consequently, we introduce the mask to guide the learning of attention map. Intuitively, instead of directly using interpolation to resize the mask to the same size of $\bm{f}_{img}$, we propose to use average pooling and normalization to process the mask since it may be more noise tolerant. The guidance is given by a Mean Squared Error (MSE) loss:
\begin{equation}
 L_{att}=\frac{1}{H\times W}\sum_{i=1}^{H}\sum_{j=1}^{W}||M_{i,j}-\bm{Am}_{i,j}||,
\end{equation}
where $i$ and $j$ are the coordinates in the resized mask and the attention map, and $M$ is the resized mask.\par
Eventually, the problem turns to where to add the proposed mask-guided module. It should be emphasized again that training such a ultra-fine-grained classification model from scratch is unrealistic due to only few data available. To obtain an acceptable precision, we have to freeze the first few layers of the pre-trained networks to guarantee the powerful low-level feature extraction capability, which means that we are unable to add this module into the middle level image features. Alternatively, as shown in Figure 2, we integrate the mask-guided module after the last Conv layer. Even though this approach affects the performance of mask-guided module, it still can provide appropriate information to aid the learning of the attention map. \par
We adopt the Cross Entropy loss $L_{ce}$ as the loss of the classification task. The final object function is:
\begin{equation}
 L=L_{ce}+\tau L_{att}
\end{equation}
where $\tau$ is a hyper-parameter (0.1 in our work) to strike the balance between two losses.
\section{Experimental results}

\subsection{Datasets}
\textbf{SoyCultivarVein Dataset}. The SoyCultivarVein dataset \cite{yu2019multiscale} is a public leaf dataset consisting of 200 categories (cultivars) with 6 samples per category. Hence, there are $200\times 6=1200$ images in this dataset. All the 200 categories come from the same specie, making this dataset a highly challenging dataset due to the significant similarity between categories. The SoyCultivarVein dataset provides both the image-level and pixel-level labels.  We split the dataset into training set and test set with a ratio of 2:1 in classification task.\par
\noindent
\textbf{HainanLeaf Dataset}. Provided by \cite{yu2019contour}, HainanLeaf dataset contains 100 images with 20 categories. Hence, it is also a small-scale fine-grained dataset with 5 samples for each category. Similar to SoyCultivarVein, the dataset is divided into training set and test set with a ratio of 3:2 for fine-grained classification.\par
\noindent
\textbf{BtfPIS dataset}. The butterfly patchy image structure dataset (BtfPIS) \cite{wang2009learning} has 500 images of ten categories (50 for each category). 100 images are used to train the model and the remaining 400 images are for testing in fine-grained classification.

\subsection{Implementation Details}
All the backbone networks adopted in this paper are pre-trained on ILSVRC 2012 dataset \cite{krizhevsky2017imagenet} as the feature extractor. As stated in previous sections, we first train a semantic segmentation model to generate the mask for each image. Since no mask information is available on the HainanLeaf dataset, to implement our idea, we manually label the veins of leaves in 60 images on HainanLeaf dataset, and transfer the learning from SoyCultivarVein to HainanLeaf by fine-tuning the model trained on SoyCultivarVein according to these manually labelled images. In fine-grained patchy image classification, random cropping, random horizontal flipping and random erasing are applied to augment the training images. The generated masks follow the same data augmentation methods as the training images and are not used in the test phase. We select the SGD optimizer with a momentum of 0.9 to optimize the parameters of the models. The initial learning rate is set to 0.05 for DenseNet161, 0.005 for VGG19 and 0.01 to other methods, and decays via the cosine annealing algorithm. 

\subsection{Semantic Segmentation Results}
Here, we show the semantic segmentation results on the SoyCultivarVein, HainanLeaf and BtfPIS datasets. It can be seen from Table 1 that HMSANet achieves a mIoU of 95.93 on BtfPIS dataset, while segmented results are not very successful on SoyCultivarVein with a mIoU of 52.23. Besides, HainanLeaf demonstrates a mIoU of less than 42. Some false positives on HainanLeaf may be caused by subjective factors, e.g., mislabeling the main vein and the second-order vein. Though the performance of our MGA module depends on the segmentation results, our methods gain the best performance on some datasets with low-quality predicted masks, indicating that this reliance is not very strict (see Table 2). Some results are visualized in Figure 1.
\vspace{-0.5cm}
\begin{table}[!h]
\centering
\caption{The semantic segmentation results (mIoU) on the test set of SoyCultivarVein (SCV), the test set of HainanLeaf and the test set of BtfPIS datasets.}
\begin{tabular}{cccc}
\toprule  
& SCV& HainanLeaf& BtfPIS\\
\midrule  
mIoU&  52.53&  41.31& 95.93\\
\bottomrule 
\end{tabular}
\end{table}

\subsection{Classification Performance}
This section experimentally compares our proposed method to some state-of-the-art classification networks on the test set of three datasets as shown in Table 2. The last line is our proposed method which takes the DenseNet161 as the backbone. Obviously, our method achieves the best accuracy of 45.75\% on the test set of SoyCultivarVein and 95\% on the test test of BtfPIS, outperforming the second best performed method by 2.75\% and 3.5\%, respectively. Besides, the best performance also can be seen on our method on the test set of HainanLeaf dataset with an accuracy of 90\%, while the DenseNet161 and MobileNetv2 obtain the same score. There may be two reasons leading to no remarkable improvement on HainanLeaf dataset. (1) The low-quality of generated masks can not provide reliable information to supervise the learning of the attention map; (2) The performance may be saturated, and hence attention mechanism can hardly contribute to the classification of these hard examples.
\vspace{-0.5cm}
\begin{table}[!h]
\caption{The accuracy (\%) of different methods on the test set of SoyCultivarVein (SCV), the test set of HainanLeaf (HL) and the test set of BtfPIS datasets. }
\centering
\begin{tabular}{cccc}
\toprule  
Method& SCV& HL& BtfPIS \\
\midrule  
VGG19 \cite{simonyan2014very}&  13.50&   77.50& 89.50\\
InceptionV3 \cite{szegedy2016rethinking}& 16.50& 87.50& 55.25 \\
ResNet50 \cite{he2016deep}& 33.50& 87.50& 89.75\\
MobileNetV2 \cite{sandler2018mobilenetv2}& 39.00& 90.00& 91.50\\
DenseNet161 \cite{huang2017densely}&  43.00&  90.00& 69.75\\
\bottomrule 
Ours&  45.75&  90.00& 95.00\\
\bottomrule 
\end{tabular}
\end{table}

\subsection{Comparison to Handcrafted Feature Based Methods}
Table 3 summarizes the comparison of our proposed method to state-of-the-art handcrafted feature based methods on the test set of SoyCultivarVein dataset. We follow the same settings in \cite{yu2020patchy} to split the dataset into training and test sets. Our method achieves an accuracy of 32.2\%, surpassing the best-performed handcrafted feature based approach by a large margin (+3.7\%).
\vspace{-0.5cm}
\begin{table}[!h]
\caption{The comparison (\%) of our method to state-of-the-art handcrafted feature based approaches on the test set of SoyCultivarVein. }
\centering
\begin{tabular}{cc}
\toprule  
Method& SoyCultivarVein \\
\midrule  
MDM \cite{hu2012multiscale}&  13.4\\
DBCSR(uni) \cite{demisse2017deformation}& 14.1\\
MORT \cite{yu2020patchy}& 28.5\\
\bottomrule 
Ours&  32.2\\
\bottomrule 
\end{tabular}
\end{table}

\subsection{Ablation studies}
We conduct the ablation study of the mask-guided attention module (MGA) on the state-of-the-art backbone network DenseNet161 on the test set of SoyCultivarVein and the test set of BtfPIS. As shown in Table 4, the proposed MGA module improves the accuracy by 2.25\% and 2\% on the SoyCultivarVein and the BtfPIS datasets compared with the unsupervised attention methods, respectively, which confirms our theory. Note that no improvement can be seen when adding the attention mechanism to the Densent161 on SoyCultivarVein, demonstrating the difficulty in further improving the accuracy on SoyCultivarVein.
\vspace{-0.5cm}
\begin{table}[!h]
\caption{The ablation study results (\%) of mask-guided attention module on the test set of SoyCultivarVein (SCV) and the test set of BtfPIS.}
\centering
\begin{tabular}{ccccc}
\toprule  
Baseline& Attention& MGA & SCV& BtfPIS\\
\midrule  
DenseNet161&   & &43.50 &69.75\\
DenseNet161& $\surd$&  & 43.50& 93.00\\
DenseNet161& $\surd$&  $\surd$& 45.75 & 95.00\\
\bottomrule 
\end{tabular}
\end{table}

\section{conclusion}
This paper proposes a novel mask-guided attention approach for the fine-grained patchy image classification. There are two core problems hindering the performance of CNN based methods on fine-grained patchy image classification, ultra-fine-grained inter-category differences and few training data. To handle these problems, we propose to utilize mask signal generated by the pre-trained semantic segmentation model to help classifier focus on the most informative parts in images. This is achieved by considering the mask as the ground truth of the attention map. Experimental results demonstrate that this approach achieves the best performance on three publicly available patchy image datasets compared with the state-of-the-art methods, and yields a significant improvement against the unsupervised attention manner.  

\clearpage
\printbibliography  
\end{document}